\documentclass{article}
\usepackage{ijcai22}

\usepackage{times}
\usepackage{soul}
\usepackage{url}
\usepackage[hidelinks]{hyperref}
\usepackage[utf8]{inputenc}
\usepackage[small]{caption}
\usepackage{graphicx}
\usepackage{amsmath}
\usepackage{amsthm}
\usepackage{booktabs}
\usepackage{algorithm}
\usepackage{algorithmic}
\usepackage{multirow}
\usepackage{makecell}
\usepackage{xcolor}

\usepackage{amsmath}
\usepackage{amssymb}
\usepackage{arydshln}

\newcommand{\newcite}[1]{\citeauthor{#1}~\shortcite{#1}}
\newcommand{\CalD}{\mathcal{D}}

\usepackage{xspace}
\newcommand{\sys}{{\textsc{All}}\xspace}

\title{A Survey on Dialogue Summarization: Recent Advances and New Frontiers}

\author{
	Xiachong Feng
	\and
	Xiaocheng Feng\thanks{~Corresponding author.} 
	\and
	Bing Qin
	\affiliations
	Harbin Institute of Technology, China \\
	\emails
\{xiachongfeng, xcfeng, bqin\}@ir.hit.edu.cn
}

\begin{document}

\maketitle

\begin{abstract}
Dialogue summarization aims to condense the original dialogue into a shorter version covering salient information, which is a crucial way to reduce dialogue data overload.
Recently, the promising achievements in both dialogue systems and natural language generation techniques drastically lead this task to a new landscape, which results in significant research attentions. However, there still remains a lack of a comprehensive survey for this task. To this end, we take the first step and present a thorough review of this research field carefully and widely.
In detail, we systematically organize the current works according to the characteristics of each domain, covering meeting, chat, email thread, customer service and medical dialogue.
Additionally, we provide an overview of publicly available research datasets as well as organize two leaderboards under unified metrics.
Furthermore, we discuss some future directions, including faithfulness, multi-modal, multi-domain and multi-lingual dialogue summarization, and give our thoughts respectively. We hope that this first survey of dialogue summarization can provide the community with a quick access and a general picture to this task and motivate future researches.
\end{abstract}

\section{Introduction}
Dialogue summarization aims to distill the most important information from a dialogue into a shorter passage, which can help people quickly capture the highlights of a semi-structured and multi-participant dialogue without reviewing the complex dialogue context \cite{emnlp2020-mvbart-chen}. With the development of communication technology and the ravage of COVID-19, different types of dialogues have emerged as an important way for information exchange. Therefore, there is an urgent need for summarization techniques to save people from large amounts of dialogue data.  

Conventional works mainly focus on single-participant document summarization, such as news and scientific papers \cite{pgn}.
Thanks to the neural models, especially the sophisticated pre-trained language models, which have advanced these tasks significantly \cite{bart}.
Despite the success of single-participant document summarization, these methods can not be easily transferred to the multi-participant dialogue summarization.
Firstly, the dialogue contains multiple participants, inherent topic drifts, frequent coreferences, diverse interactive signals and domain terminologies \cite{ijcai2021-feng}. All of these characteristics make dialogue a hard-to-model data type.
Secondly, in terms of different domains, the above characteristics further pose domain-specific challenges to summarization models, e.g., \textit{How to model long meeting transcripts} \cite{emnlp2020-hmnet-zhu}.
Thirdly, compared with widely used document summarization benchmarks, collecting labeled dialogue-summary paired data is highly-costing or even intractable \cite{chen-yang-2021-simple}. 
To mitigate these challenges, researchers draw on successful experiences from the study of dialogue systems and natural language generation techniques and put their efforts on solving this challenging task, which result in nearly 100 papers covering various domains being published over the past 5 years.

To review the current progress and help new researchers get into the field quickly, we present this first survey for dialogue summarization.
As the preliminary, we quickly overview the recent progress in general summarization and capture several key time points and key techniques, this serves as a strong background before we dive into the dialogue summarization (see \S\ref{sec:bg}).
As the core content, we summarize existing works according to the domain of dialogue, mainly covering the meeting, chat, email thread, customer service and medical dialogue. 
For each type of dialogue, we thoroughly go through related research works, organize them according to their unique challenges and provide suggestions for future works (see \S\ref{sec:tax}). 
For example, we focus on two main streams of works for chat summarization including interaction and participant modeling \cite{CoreferenceAwareDS,liu-chen-2021-controllable}.
In terms of customer service, we organize related works from two perspectives, one is inherent topic modeling \cite{cs-kdd}, the other is task-oriented-specific information integration \cite{todsum}.
Besides, we provide an overview of publicly available research datasets (see Table \ref{tab:datasets}). Especially for meeting and chat summarization, we also carefully organize leaderboards under the unified evaluation metric by collecting reported results from published literatures and re-evaluating official outputs (see Table \ref{tab:ami_bench} and Table \ref{tab:samsum_bench}).
Based on the analyses of existing works, we present several research directions, including faithfulness in dialogue summarization, multi-modal, multi-domain and multi-lingual dialogue summarization (see \S\ref{sec:new}). All of these frontiers not only pose new research challenges but also meet actual application needs and fit in with real-world scenarios.

To sum up, our contributions are as follows: 
\begin{itemize}
    \item We are the first to present a comprehensive survey for the dialogue summarization task.
    \item We thoroughly summarize existing works according to different types of dialogues and carefully organize leaderboards under the unified evaluation metric.
    \item We discuss some new frontiers and highlight their challenges to motivate future researches.
\end{itemize}

\section{Background}\label{sec:bg}
In this section, we give an overview on the summarization task, then describe the commonly used evaluation metrics. 

\subsection{Overview of Summarization}
Automatic summarization is a fundamental task in natural language processing and has been continuously
studied for decades \cite{old1}.
It aims to condense the original input into a shorter version covering salient information, which can help people quickly grasp the core content without diving into the details. 
It is mainly divided into two paradigms: \textit{extractive} and \textit{abstractive}. 
Extractive methods select vital sentences as the summary, which is more accurate and faithful, while abstractive methods generate the summary using novel words, which improves the conciseness and fluency of the summary. 
Previous works adopt machine learning algorithms to perform extractive summarization \cite{textrank}. 
With sophisticated neural architectures, data-driven approaches have made much progress in both two paradigms. 
Especially for abstractive methods, sequence-to-sequence learning combined with attention mechanism is adopted as the backbone architecture for solving this task \cite{pgn}.
Recently, with the great success of pre-trained models in a wide range of natural language processing tasks, these models also become the {\it de facto} way for summary generation and have achieved many state-of-the-art results \cite{bart}. 

\subsection{Evaluation Metrics}
ROUGE \cite{rouge} is conventionally adopted as the standard metric for evaluating summarization tasks, which mainly involves the F1 scores for ROUGE-1, ROUGE-2, and ROUGE-L that measure the word overlap, bi-gram overlap and longest common sequence between the ground truth and the generated summary respectively. 

\begin{table}[t]
\small
\centering
        \scalebox{0.90}{\begin{tabular}{l|c|c}
            \hline
            \textbf{Name} & \textbf{Domain} & \textbf{Language} \\
            \hline
            \hline
            ICSI \cite{icsi}  & \multirow{3}{*}{Meeting}  & English \\
            AMI \cite{ami}  &  & English \\
            QMSum \cite{qmsum} &  & English \\
            \hdashline[1pt/3pt]
            SAMSum \cite{emnlp2019-samsum-gliwa}  & \multirow{2}{*}{Chat} & English \\
            GupShup \cite{gupshup}    &  & Code-Mix \\
            \hdashline[1pt/3pt]
            CSDS \cite{csds} & \multirow{3}{*}{\makecell{Customer \\ Service} }  & Chinese \\
            TODSum \cite{todsum} &   & English  \\
            TWEETSUMM \cite{tweetsumm}  &  & English \\
            \hdashline[1pt/3pt]
            CRD3 \cite{crd3}   & TV Show & English \\
            \cite{coling2020-medical-song}   & Medical & Chinese \\
            SumTitles \cite{coling2020-titles-malykh}   & Movie & English \\
            \textsc{MediaSum} \cite{naacl2021-media-zhu} & Interview & English \\
            \textsc{DialogSum} \cite{acl2021-dialsum-chen} & Spoken & English \\
            $\textsc{EmailSum}$ \cite{emailsum} & Email & English \\
            ForumSum \cite{forumsum} & Forum & English \\
            \hdashline[1pt/3pt]
            ConvoSumm \cite{convosumm}   & Mix & English \\
            \hline
        \end{tabular}}
\caption{Major datasets for dialogue summarization.} \label{tab:datasets}
\end{table}

\section{Taxonomy}\label{sec:tax}
In this section, we describe the taxonomy of dialogue summarization according to the domain of input dialogue, including meeting, chat, email thread, customer service and medical dialogue. Table \ref{tab:datasets} lists currently available datasets for these dialogue summarization researches.

\begin{table*}[t]
\small
\centering
        \begin{tabular}{l||ccc||ccc}
            \hline
            &  \multicolumn{3}{c||}{\textbf{AMI}} & \multicolumn{3}{c}{\textbf{ICSI}} \\

            \textbf{Model} & \textbf{ROUGE-1} & \textbf{ROUGE-2} & \textbf{ROUGE-L} & \textbf{ROUGE-1} & \textbf{ROUGE-2} & \textbf{ROUGE-L}  \\
            \hline
            \hline
            \multicolumn{7}{c}{\it Extractive Methods} \\
            \hline
            TextRank \cite{textrank} &35.19 &6.13 &15.70 &30.72 &4.69 &12.97  \\
            SummaRunner \cite{aaai2017-summarurnn-nallapati} & 30.98 &5.54 &13.91 &27.60 &3.70 &12.52 \\
            \hline
            \hline
            \multicolumn{7}{c}{\it Abstractive Methods} \\
            \hline
            UNS \cite{acl2018-shang} &37.86 &7.84 &13.72 & 31.73 & 5.14 & 14.50 \\
            PGN \cite{pgn} &42.60 &14.01 &22.62 & 35.89 & 6.92 & 15.67 \\
            Sentence-Gated \cite{goo} & 49.29 & 19.31 & 24.82 & 39.37 & 9.57 & 17.17 \\
            TopicSeg \cite{acl2019-limanling} & 51.53 &12.23 &25.47  & - & - & - \\
            TopicSeg+VFOA \cite{acl2019-limanling} & 53.29 &13.51 &26.90  & - & - & - \\
            HMNet \cite{emnlp2020-hmnet-zhu} & 52.36 & 18.63 &24.00 & 45.97 & 10.14 & 18.54 \\
            PGN($\CalD_{\sys}$) \cite{acl2021-gpt-feng} &50.91 &17.75  &24.59 & - & - & - \\
            \textsc{DdaMS} \cite{ijcai2021-feng} &51.42 &20.99 &24.89 &39.66 &10.09 &17.53 \\
            \textsc{DdaMS}+\textsc{DdaDA} \cite{ijcai2021-feng} &53.15 &22.32 &25.67 &40.41 &11.02 &19.18 \\
            \hline
            \hline
            \multicolumn{7}{c}{\it Pre-trained Language Model-based Methods} \\
            \hline
            Longformer-BART \cite{convosumm} &54.81  &20.83  &25.98  & 43.40 & 12.19 & 19.29 \\
            Longformer-BART-arg \cite{convosumm} & 55.27 & 20.89 & 24.94 & 44.51 & 11.80 & 19.19 \\
            \textsc{Dialog}LM \cite{DialogLM} & 53.72  & 19.61  & 51.83* & 49.56  & 12.53  & 47.08* \\
            \hline
        \end{tabular}
\caption{Leaderboard of meeting summarization on AMI \protect\cite{ami} and ICSI \protect\cite{icsi} datasets. We adopt reported results from published literatures \protect\cite{ijcai2021-feng} and corresponding publications. The results of Longformer \protect\cite{convosumm} are obtained by evaluating the output files provided by the author. Results with *
indicate that ROUGE-L is calculated with sentence splitting.} \label{tab:ami_bench}
\end{table*}

\subsection{Meeting Summarization}\label{sec:meeting}
Meeting plays an essential part in our daily life. 
Especially due to the spread of COVID-19 worldwide, people are more dependent on online meetings to share information and collaborate with others.
Accordingly, meeting summaries, aka meeting minutes could be of great value for both participants and non-participants to quickly grasp the main meeting ideas. 
Thanks to the earlier publicly available datasets AMI \cite{ami} and ICSI \cite{icsi}, meeting summarization has attracted extensive research attentions. 

Precedent works focus on extractive meeting summarization. They adopt various features to detect important utterances, such key phrases, topics and speaker characteristics.
However, due to the multi-participants nature, information is scattered and incoherent in the meeting, which makes the extractive methods unsuitable for meeting summarization. 
Therefore, recent years witness a growing trend of abstractive meeting summarization methods \cite{acl2018-shang}.

With the development of neural networks, many works have explored the application of deep learning in meeting the summarization task and have achieved remarkable success \cite{emnlp2020-hmnet-zhu}.
Although deep learning-based methods have strong modeling abilities, taking only literal information into consideration is not sufficient. 
This is because there are diverse interactive signals among meeting utterances and the long meeting transcripts further pose challenges to traditional sequence-to-sequence models.
To this end, some works devote efforts to incorporate auxiliary information for better modeling meetings, such as dialogue discourse \cite{ijcai2021-feng}, dialogue acts \cite{goo} and domain terminologies \cite{coling2020-term-koay}.
Besides, several strategies are carefully devised to handle long meeting transcripts, including hierarchical modeling strategy \cite{emnlp2020-hmnet-zhu}, sliding window strategy \cite{Koay2021ASA}, retrieve-then-summarize strategy \cite{zhang-etal-2021-exploratory-study} and pre-training strategy \cite{DialogLM}.

Instead of summarizing the whole meeting, generating meeting summaries of a particular aspect, such as decisions, actions, ideas and hypotheses, could also address specific needs.
Recently, \newcite{qmsum} propose the query-based meeting summarization task, which aims to summarize the specific part of the meeting according to the given query.

In addition to multi-party characteristics, meeting summarization has also been explored under the multi-modal setting. Meetings can include various types of non-verbal information that is displayed by the participants, such as audio, visual and motion features.
These features may be useful for detecting important utterances in a meeting.
Therefore, a majority of works study both the extractive and abstractive multi-modal meeting summarization problem by fusing verbal and non-verbal features to enrich the representation of the utterance \cite{acl2019-limanling}.

\paragraph{Leaderboard:}
To unify this research direction, we systematically present a comprehensive leaderboard for two widely used meeting summarization datasets: the AMI and ICSI datasets, using \textit{pyrouge} package\footnote{https://pypi.org/project/pyrouge/}, as shown in Table \ref{tab:ami_bench}.

\paragraph{Highlight:} 
Meetings always involve several participants with specific roles. Thus, it is necessary to model such distinctive role characteristics. 
Besides, the long transcripts also need the model to be capable of handling long sequences. 
Furthermore, the audio-visual recordings of meetings provide the opportunity for using multi-modal information. 
However, it is a double-edged sword. The error rate of automatic speech recognition systems and vision tools also pose challenges to the current models, which requires them to be more robust.

\subsection{Chat Summarization}\label{sec:chat}
Online chat applications have become an indispensable way for people to communicate with each other, which has led to people being overwhelmed by massive amounts of chat information.
Such complex dialogue context poses a challenge to the new chat participant, since he or she may be unable to quickly review the main idea of the dialogue.
Therefore, summarizing chats becomes a new trending direction.

\newcite{emnlp2019-samsum-gliwa} introduce the first high-quality and manually annotated chat summarization corpus, namely, SAMSum, and conduct various baseline experiments, which rapidly sparks this research direction. Afterward, \newcite{emnlp2020-mvbart-chen} take the first step and propose a multi-view dialogue summarizer by introducing both topic segments and conversational stages. More importantly, they conduct a comprehensive study for the challenges in this task, revealing the importance of dialogue modeling for the dialogue summarization, which points out the direction for future researchers.

Roughly speaking, the majority of current works put much emphasis on two aspects: \textit{dialogue interaction modeling} and \textit{dialogue participant modeling}, which are in line with the prominent characteristics of conversational data.

To model the interaction, graph modeling strategies combined with additional features are widely adopted.
\newcite{coling2020-topic-zhao} utilize fine-grained topic words as bridges between utterances to construct a topic-word guided dialogue graph.
\newcite{naacl2021-discourse-chen} consider the inter-utterance dialogue discourse structure and intra-utterance action triples to explicitly model the interaction. 
\newcite{ccl2021-know-feng} view commonsense knowledge as cognitive interactive signals behind different utterances and shows the effectiveness of the integration of knowledge and heterogeneity modeling for different types of data.
\newcite{CoreferenceAwareDS} explicitly incorporate coreference information in dialogue summarization models. It is worth noting that they conduct data postprocessing to reduce incorrect coreference assignments caused by document coreference resolution model.

To model the participants, \newcite{speaker} implicitly model complex relationships among participants and their relative personal pronouns via speaker-aware self-attention mechanism. 
From another perspective, \newcite{frost} and \newcite{liu-chen-2021-controllable} explicitly adopt the guided summarization framework and introduce the participant information into the coarse-to-fine generation procedure, in which the final dialogue summary is controlled by a precedent, such as a sketch or named entities. 

As shown in the above works, current dialogue summarization systems usually encode the text with additional information.
However, these annotations are usually obtained via open-domain toolkits, which are not suitable for dialogues, or require manual annotations, which are labor-consuming.
Therefore, \newcite{acl2021-gpt-feng} present an unsupervised DialoGPT annotator, which can perform three dialogue-specific annotation tasks, including keywords extraction, redundancy detection and topic segmentation. 

Despite the encouraging results reported, current models still suffer from the data-insufficient problem. Accordingly, some researchers study this task in the low-resource regime.
\newcite{augfords} innovatively explore the summary-to-dialogue generation problem and verify the augmented dialogue-summary pairs can do good to dialogue summarization.
\newcite{chen-yang-2021-simple} propose three conversational data augmentation methods to enrich the data, including random swapping or deletion utterances, dialogue-acts-guided utterance insertion and conditional-generation-based utterance substitution.

\paragraph{Leaderboard:} 
Previous works have already achieved remarkable success on the SAMSum dataset \cite{emnlp2019-samsum-gliwa}. 
However, due to the different versions of ROUGE evaluation package, there lacks benchmark results unifying all the scores. To this end, we present benchmark results using \textit{py-rouge} package\footnote{https://pypi.org/project/py-rouge/}. The results are shown in Table \ref{tab:samsum_bench}.

\paragraph{Highlight:}
Thanks to the pre-trained language models, current methods are skilled at transforming the original chat into a simple summary realization. However, they still have difficulty selecting the important parts and tend to generate hallucinations. In the future, powerful chat modeling strategies and reasoning abilities should be explored for this task, and more low-resource settings should be considered.

\begin{table}[t]
\small
\centering
        \begin{tabular}{l|ccc}
            \hline
            \textbf{Model} & \textbf{R-1} & \textbf{R-2} & \textbf{R-L}  \\
            \hline
            \hline
            \multicolumn{4}{c}{\it Extractive Methods} \\
            \hline
            LONGEST-3 &32.46 &10.27 &29.92 \\
            TextRank \cite{textrank} &29.27 &8.02 &28.78 \\
            \hline
            \hline
            \multicolumn{4}{c}{\it Abstractive Methods} \\
            \hline
            Transformer \cite{transformer} &36.62 &11.18 &33.06 \\
            PGN \cite{pgn} & 40.08 & 15.28 & 36.63 \\
            D-HGN \cite{ccl2021-know-feng} &42.03  &18.07  &39.56   \\
            TGDGA \cite{coling2020-topic-zhao} &43.11 &19.15 &40.49 \\
            \hline
            \hline
            \multicolumn{4}{c}{\it Pre-trained Language Model-based Methods} \\
            \hline
            DialoGPT \cite{dialogpt} &39.77 &16.58 &38.42 \\
            \hdashline[1pt/3pt]
            UniLM \cite{unilm} & 47.85 & 24.23 & 46.67 \\
            BART \cite{bart} &52.98 &27.67 &49.06 \\
            \hdashline[1pt/3pt] 
            S-BART \cite{naacl2021-discourse-chen} & 50.70 & 25.50 & 48.08 \\
            \textsc{Frost} \cite{frost} & 51.86 & 27.67 & 47.52 \\
            CODS \cite{cods} & 52.65 & 27.84 & 50.79 \\
            MV-BART \cite{emnlp2020-mvbart-chen} &54.05 &28.56  &50.57 \\
            BART($\CalD_{\sys}$) \cite{acl2021-gpt-feng} &53.70 &28.79 &50.81 \\
            Coref-ATTN \cite{CoreferenceAwareDS} & 53.93 & 28.58 & 50.39 \\
            \hdashline[1pt/3pt] 
            Entity-Plan \cite{liu-chen-2021-controllable}$^{\dagger}$ & $\text{56.53}$ & $\text{32.40}$ & $\text{54.92}$ \\
            \hline
        \end{tabular}
\caption{Leaderboard of chat summarization on the SAMSum dataset \protect\cite{emnlp2019-samsum-gliwa}, where ``R'' is short for ``ROUGE''. We mainly adopt results from corresponding publications. Besides, the results of S-BART, MV-BART, Coref-ATTN and Entity-Plan are obtained by evaluating output files provided by the author. $\dagger$ indicates the model obtains these results with the help of golden summaries.}
\label{tab:samsum_bench}
\end{table}

\subsection{Email Threads Summarization} \label{sec:email}
Email thread is an asynchronous multi-party communication consisting of a coherent exchange of email messages among several participants, which is widely used in the enterprise, academic and work settings. 
Compare with other types of dialogue, email has some unique characteristics. 
Firstly, it associates with the meta-data, including sender, receiver, main body and signature.
Secondly, the email message always represents the intent of the sender, contains action items and may use quote to highlight the important part. 
Thirdly, unlike face-to-face spoken dialogue, replies in the email do not happen immediately. Such asynchronous nature may result in messages containing long sentences.
To deal with email overload, email service providers seek for efficient summarization techniques to improve the user experience.

Major efforts lie on email thread summarization. Pioneer works present publicly available datasets to facilitate this task. 
\newcite{email-data-1} collect 39 email threads from Enron email dataset and annotate them with extractive summaries.
They propose an email fragment quotation graph based on the occurrence of clue words and conduct extractive summarization. 
Notably, quotation plays an important role in the email that can directly highlight the salient part of the previous email.
To enrich the annotation, 
\newcite{email-data-2} collect 40 email threads from W3C email dataset and annotate them with both abstractive and extractive summaries along with meta sentences, subjectivity and speech acts.
\newcite{email-data-3} collect 107 email threads from Enron email dataset and annotate them with extractive and abstractive summaries combined with key phrases. 
Recently, \newcite{emailsum} present \textsc{EmailSum}, which contains 2549 email threads collected from Avocado Research Email Collection associated with human-written short and long abstractive summaries. This large-scale and high-quality dataset provides opportunities to data-hungry neural models.

In light of emails always being used for workflow organization and task tracking, some works explore action-focused email summarization, aka TO-DO item generation, which can help users with task management over emails.
\newcite{email-todo-new} propose a Smart TO-DO system, which first detects commitment sentences and then generates to-do items using sequence-to-sequence models. 

\paragraph{Highlight:} 
Email is a specific genre of dialogue, which aims to organize the workflow.
Therefore, an email frequently proposes requests, makes commitments and contains action items, which make the email intent understanding of vital importance. 
Future works should pay more attention to understanding the fine-grained action items in the email and the coarse-grained intent of the entire email.
Besides, better use of quotations can yield significant benefits.

\subsection{Customer Service Summarization}\label{sec:cs}
Customer service is the direct one-on-one interaction between a customer and an agent, which frequently happens before and after the consumer behavior.
Thus, it is important for growing business. 
To make the customer service more effective, automatic summarization is one way, which can provide the agent with quick solutions according to the previous condensed summary.
Therefore, customer service summarization gains a lot of research interest in recent years.

On the one hand, participants in customer service have strong intent and clear motivations to address issues, which makes the customer service inherently logical and surrounds specific topics. To this end, some works explore topic modeling for this task.
\newcite{cs-kdd} employ a coarse-to-fine generation framework, which first generates a sequence of key points (topics) to indicate the logic of the dialogue and then realize the detailed summary. For example, a key point sequence can be {\it question$\rightarrow$solution$\rightarrow$user approval$\rightarrow$end}, which clearly shows the evolution of the dialogue. 
Instead of using explicitly pre-defined topics, \newcite{cs-supervised-zou} draw support from neural topic modeling and propose a multi-role topic modeling mechanism to explore implicitly topics. 
To alleviate data-insufficient problems, \newcite{cs-unsupervised-zou} propose an unsupervised framework called RankAE, in which  topic utterances are first selected  according to centrality and diversity simultaneously, and the denoising auto-encoder is further employed to produce final summaries.

On the other hand, customer service is a kind of task-oriented dialogue, which contains informative entities, covers various domains and involves two distinct types of participants.
To integrate dialogue-specific information, \newcite{todsum} craft a new dataset annotated with dialogue state knowledge, which is helpful for tracking the fine-grained dialogue information flow and generating faithful summaries.
Since participants in customer service play distinct roles, in addition to the overall summary
for the whole dialogue, \newcite{cs-tat} propose an unsupervised framework based on variational auto-encoder to generate summaries for the customer and the agent respectively. 
\cite{csds} directly propose CSDS  datasets annotated with role-oriented summaries to acquire different speakers’ viewpoints.

\paragraph{Highlight:} 
Customer service aims to address the questions raised by agents. Therefore, it naturally has strong motivations, which makes the dialogue have a specific way of evolution following the interaction between two participants with distinctive characteristics: the customer and the agent. Thus, modeling participant roles, evolution chains and inherent topics are important for this task. Besides, some fine-grained information also should be taken into consideration to ensure faithfulness, such as slots, states and intents.

\subsection{Medical Dialogue Summarization}\label{sec:medical}
Medical dialogues happen between patients and doctors. During this process, doctors are required to record a digital version of a patient’s health records, namely electronic health records (EHR), which leads to both patient dis-satisfaction and clinician burnout.
To mitigate the above challenge, medical dialogue summarization is coming to the rescue. 

From a coarse-grained perspective, a medical dialogue can be divided into several coherent segments according to different criteria.
\newcite{medical-topic-1} specify the dialogue topics according to the symptoms, such as headache and cough, and design a topic-level attention mechanism to make the decoder focus on one symptom when generating one summary sentence.
\newcite{medical-topic-2} instead choose EHR categories to label each segment, such as family history and medical history.
Specifically, \newcite{soap} name the medical dialogue summary {\it SOAP note}, which stands for \underline{S}ubjective information reported by the patient; \underline{O}bjective observations; \underline{A}ssessments made by the doctor; and a \underline{P}lan for future care, including diagnostic tests and treatments. 

From a fine-grained perspective, several medical dialogue characteristics should be handled carefully.
Firstly, question-answer pairs are the major discourse in medical dialogue and negations scattered in different utterances are notable parts. 
To this end, \newcite{drsummarize} encourage the model to focus on negation words via negation word attention and explicitly employ a gate mechanism to generate the \textit{[NO]} word.
Secondly, medical terminologies play an essential part in medical dialogues.
\newcite{drsummarize} leverage the compendium of medical concepts, known as unified medical language systems to identify the presence of terminologies and further use an indicator vector to influence the attention distribution.
Thirdly, the medical dialogue summary mainly describes core items and concepts in the dialogue, therefore, the summarization methods should bias towards extractive methods while keeping the advantages of abstractive methods. \newcite{enarvi-etal-2020-generating} and \newcite{drsummarize} both enhance the copy mechanism to facilitate copying from the input.

\paragraph{Highlight:} 
Medical dialogue summarization mainly aims at helping doctors to quickly finish electronic health records and the medical dialogue summary should be more faithful rather than creative. Therefore, extractive methods combined with simple abstractive methods are preferred. 
The topic information can serve as a guideline for generating semi-structured summaries.
Besides, terminologies and negations in the medical dialogue should be handled carefully.

\section{New Frontiers}\label{sec:new}
Section \ref{sec:tax} mainly introduces prominent achievements in different domains respectively.
In this section, we will discuss some new frontiers which meet actual application needs and fit in with real-world scenarios.

\subsection{Faithfulness in Dialogue Summarization}
Even though current state-of-the-art summarization systems have already made great progress, they still suffer from the factual inconsistency problem, which distorts or fabricates the factual information in the article and is also known as hallucinations \cite{tang2021investigating}. \newcite{Tang2021CONFITTF} systematically study the taxonomy of factuality errors for dialogue summarization,  which includes the following 8 error types: \textit{Missing Information, Redundant Information, Circumstantial Error, Wrong Reference Error, Negation Error, Object Error, Tense Error} and \textit{Modality Error}. 
Specifically, the last five types of errors notoriously tend to appear in dialogue summaries, which largely hinder the application of dialogue summarization systems. 

To remedy these issues, future works need specific designs target for the above errors. Importantly, fine-grained dialogue-specific features need to be incorporated into the summarization model, such as personal pronoun information, coreference information and tense information. On the one hand, these features can implicitly alleviate the difficulty of dialogue understanding. On the other hand, some features can directly serve as the explicitly extracted information to help final summary generation.

\subsection{Multi-modal Dialogue Summarization}
Dialogues tend to occur in multi-modal situations, such as audio-visual recordings of meetings.
Besides verbal information, non-verbal information can either supplement existing information or provide new information, which effectively enriches the representation of purely textual dialogues. 
According to whether different modalities can be aligned, the types of multi-modal information can be divided into two categories: synchronous and asynchronous.

Synchronous multi-modal dialogues mainly refer to meetings, which may contain textual transcripts, prosodic audios and visual videos. 
On the one hand, taking the aligned audio and video into consideration can enhance the representation of transcripts.
On the other hand, both the audio and video can provide new insights, such as a person entering the room to join the meeting or an emotional discussion.
However, facial features and voiceprint features have already become superior privacy for individuals, which makes them hard and sensitive to be acquired. Future works can consider multi-modal meeting summarization under the federal learning framework.

Asynchronous multi-modal dialogues refer to different modalities that happen at different times. 
With the development of communication technology, multi-modal messages, such as voice messages, pictures and emoji are frequently used in chat dialogues via applications like Messenger, WhatsApp and WeChat. 
These messages provide rich information, serving as one part of the dialogue flow.
Future works should consider textual information of voice messages obtained via ASR systems, new entities provided by pictures and emotions associated with the emoji to produce meaningful summaries.

\subsection{Multi-domain Dialogue Summarization}
Multi-domain learning can mine shared information between different domains and further help the task of a specific domain, which is an effective learning method suitable for low-resource scenarios.
Thanks to diverse summarization datasets, there are already some works exploring the multi-domain learning or domain adaption for dialogue summarization \cite{AdaptSum}. 
We divide this direction into two categories: macro multi-domain learning and micro multi-domain learning. 

Macro multi-domain learning aims to use general domain summarization datasets, like news and scientific papers, to help the dialogue summarization task. 
The basis for this learning method to work is that no matter what data type they belong to, they aim to pick the core content of the original text.
However, dialogues have some unique characteristics like more coreferences and participant-related features. 
Therefore, directly using these general datasets may reduce their effectiveness.
Future works can first inject some dialogue-specific features, like replacing names with personal pronouns or transform the original general domain documents into turn-level documents at surface level, to further utilize these datasets.

Micro multi-domain learning aims to use dialogue summarization datasets to help one specific dialogue summarization task.
For example, using meeting datasets to help with email tasks. As shown in Table \ref{tab:datasets}, diverse dialogue summarization datasets covering various domains have been proposed in recent years. 
Future works can adopt meta-learning methods or rely on pre-trained language models to unify different datasets and mine common features.

\subsection{Multi-lingual Dialogue Summarization}
With the acceleration of globalization, a dialogue involving multinational participants becomes increasingly common thanks to the sophisticated instantaneous translation system. Therefore, there is an urgent need for providing people with dialogue summaries in a preferred language. However, current works overwhelmingly focused on English, while leaving other languages under exploration. We argue that the current dilemma is mainly caused by the intractable access to available multi-lingual data resources.

Firstly, future works should devote efforts to creating a suitable testbed for multi-lingual dialogue summarization. As an initial step, \newcite{gupshup} transform English utterances in the SAMSum dataset into Hindi-English utterances and study the chat summarization under the code-switched setting. 
From a higher point of view, large-scale high-quality datasets covering diverse languages should be carefully crafted.
Practically speaking, on the one hand, researchers can translate one specific dataset into different languages followed by automatic and human quality checking to get aligned datasets. On the other hand, researchers can also borrow ideas from unsupervised multi-lingual learning to utilize currently available datasets in different languages. Secondly, future works should set up systematic settings for this multi-lingual research, including \textit{one-to-one}, \textit{one-to-many}, \textit{many-to-one} and \textit{many-to-many}, in which \textit{one-to-one} setting can be further divided into \textit{mono-lingual} setting and \textit{cross-lingual} setting \cite{wang2022survey}. Thirdly, plenty of multi-lingual pre-trained language models can be explored for this task. Especially, models that have already been fine-tuned on the translation datasets may bring significant benefits

\section{Conclusion}\label{sec:con}
This article presents the first comprehensive survey on the progress of dialogue summarization carefully and widely. 
We thoroughly summarize the existing works, which cover various domains and highlight their challenges respectively. Besides, we summarize currently available datasets and organize two leaderboards.
Furthermore, we shed light on some new trends in this research field.
We hope this survey can facilitate the research of the dialogue summarization.

\appendix
\section{Acknowledgments}
We thank all the anonymous reviewers for their insightful comments. We would like to thank Alexander R. Fabbri, Jiaao Chen and Zhengyuan Liu for sharing their systems’ outputs. We would also like to thank Shiyue Zhang for her feedback on email summarization and Libo Qin for his helpful discussion. This work was supported by the National Key RD Program of China via grant 2020AAA0106502, National Natural Science Foundation of China (NSFC) via grant 61976073 and Shenzhen Foundational Research Funding (JCYJ20200109113441941).

\small
\bibliographystyle{named}
\bibliography{ijcai22}

\begin{thebibliography}{}

\bibitem[\protect\citeauthoryear{Carenini \bgroup \em et al.\egroup
  }{2007}]{email-data-1}
Giuseppe Carenini, Raymond~T. Ng, and Xiaodong Zhou.
\newblock Summarizing email conversations with clue words.
\newblock In {\em Proc. of WWW}, 2007.

\bibitem[\protect\citeauthoryear{Carletta \bgroup \em et al.\egroup
  }{2005}]{ami}
Jean Carletta, Simone Ashby, Sebastien Bourban, Mike Flynn, Mael Guillemot,
  Thomas Hain, Jaroslav Kadlec, Vasilis Karaiskos, Wessel Kraaij, Melissa
  Kronenthal, et~al.
\newblock The ami meeting corpus: A pre-announcement.
\newblock In {\em International workshop on machine learning for multimodal
  interaction}, 2005.

\bibitem[\protect\citeauthoryear{Chen and Yang}{2020}]{emnlp2020-mvbart-chen}
Jiaao Chen and Diyi Yang.
\newblock Multi-view sequence-to-sequence models with conversational structure
  for abstractive dialogue summarization.
\newblock In {\em Proc. of EMNLP}, 2020.

\bibitem[\protect\citeauthoryear{Chen and Yang}{2021a}]{chen-yang-2021-simple}
Jiaao Chen and Diyi Yang.
\newblock Simple conversational data augmentation for semi-supervised
  abstractive dialogue summarization.
\newblock In {\em Proc. of EMNLP}, 2021.

\bibitem[\protect\citeauthoryear{Chen and
  Yang}{2021b}]{naacl2021-discourse-chen}
Jiaao Chen and Diyi Yang.
\newblock Structure-aware abstractive conversation summarization via discourse
  and action graphs.
\newblock In {\em Proc. of NAACL}, 2021.

\bibitem[\protect\citeauthoryear{Chen \bgroup \em et al.\egroup
  }{2021}]{acl2021-dialsum-chen}
Yulong Chen, Yang Liu, Liang Chen, and Yue Zhang.
\newblock {D}ialog{S}um: {A} real-life scenario dialogue summarization dataset.
\newblock In {\em Proc. of Findings of ACL-IJCNLP}, 2021.

\bibitem[\protect\citeauthoryear{Dong \bgroup \em et al.\egroup }{2019}]{unilm}
Li~Dong, Nan Yang, Wenhui Wang, Furu Wei, Xiaodong Liu, Yu~Wang, Jianfeng Gao,
  Ming Zhou, and Hsiao{-}Wuen Hon.
\newblock Unified language model pre-training for natural language
  understanding and generation.
\newblock In {\em Proc. of NeuIPS}, 2019.

\bibitem[\protect\citeauthoryear{Enarvi \bgroup \em et al.\egroup
  }{2020}]{enarvi-etal-2020-generating}
Seppo Enarvi, Marilisa Amoia, Miguel Del-Agua~Teba, Brian Delaney, Frank Diehl,
  Stefan Hahn, Kristina Harris, Liam McGrath, Yue Pan, Joel Pinto, Luca Rubini,
  Miguel Ruiz, Gagandeep Singh, Fabian Stemmer, Weiyi Sun, Paul Vozila, Thomas
  Lin, and Ranjani Ramamurthy.
\newblock Generating medical reports from patient-doctor conversations using
  sequence-to-sequence models.
\newblock In {\em Proc. of the First Workshop on Natural Language Processing
  for Medical Conversations}, 2020.

\bibitem[\protect\citeauthoryear{Fabbri \bgroup \em et al.\egroup
  }{2021}]{convosumm}
Alexander Fabbri, Faiaz Rahman, Imad Rizvi, Borui Wang, Haoran Li, Yashar
  Mehdad, and Dragomir Radev.
\newblock {C}onvo{S}umm: Conversation summarization benchmark and improved
  abstractive summarization with argument mining.
\newblock In {\em Proc. of the ACL-IJCNLP}, 2021.

\bibitem[\protect\citeauthoryear{Feigenblat \bgroup \em et al.\egroup
  }{2021}]{tweetsumm}
Guy Feigenblat, Chulaka Gunasekara, Benjamin Sznajder, Sachindra Joshi, David
  Konopnicki, and Ranit Aharonov.
\newblock {TWEETSUMM} a dialog summarization dataset for customer service.
\newblock In {\em Proc. of Findings of EMNLP}, 2021.

\bibitem[\protect\citeauthoryear{Feng \bgroup \em et al.\egroup
  }{2021a}]{ccl2021-know-feng}
Xiachong Feng, Xiaocheng Feng, and Bing Qin.
\newblock Incorporating commonsense knowledge into abstractive dialogue
  summarization via heterogeneous graph networks.
\newblock In {\em Proc. of CCL}, 2021.

\bibitem[\protect\citeauthoryear{Feng \bgroup \em et al.\egroup
  }{2021b}]{ijcai2021-feng}
Xiachong Feng, Xiaocheng Feng, Bing Qin, and Xinwei Geng.
\newblock Dialogue discourse-aware graph model and data augmentation for
  meeting summarization.
\newblock In {\em Proc.of IJCAI}, 2021.

\bibitem[\protect\citeauthoryear{Feng \bgroup \em et al.\egroup
  }{2021c}]{acl2021-gpt-feng}
Xiachong Feng, Xiaocheng Feng, Libo Qin, Bing Qin, and Ting Liu.
\newblock Language model as an annotator: Exploring {D}ialo{GPT} for dialogue
  summarization.
\newblock In {\em Proc. of ACL-IJCNLP}, 2021.

\bibitem[\protect\citeauthoryear{Gliwa \bgroup \em et al.\egroup
  }{2019}]{emnlp2019-samsum-gliwa}
Bogdan Gliwa, Iwona Mochol, Maciej Biesek, and Aleksander Wawer.
\newblock {SAMS}um corpus: A human-annotated dialogue dataset for abstractive
  summarization.
\newblock In {\em Proceedings of the 2nd Workshop on New Frontiers in
  Summarization}, 2019.

\bibitem[\protect\citeauthoryear{Goo and Chen}{2018}]{goo}
Chih-Wen Goo and Yun-Nung Chen.
\newblock Abstractive dialogue summarization with sentence-gated modeling
  optimized by dialogue acts.
\newblock {\em 2018 IEEE Spoken Language Technology Workshop (SLT)}, 2018.

\bibitem[\protect\citeauthoryear{Gunasekara \bgroup \em et al.\egroup
  }{2021}]{augfords}
Chulaka Gunasekara, Guy Feigenblat, Benjamin Sznajder, Sachindra Joshi, and
  David Konopnicki.
\newblock Summary grounded conversation generation.
\newblock In {\em Proc. of Findings of ACL-IJCNLP}, 2021.

\bibitem[\protect\citeauthoryear{Janin \bgroup \em et al.\egroup }{2003}]{icsi}
Adam Janin, Don Baron, Jane Edwards, Dan Ellis, David Gelbart, Nelson Morgan,
  Barbara Peskin, Thilo Pfau, Elizabeth Shriberg, Andreas Stolcke, et~al.
\newblock The icsi meeting corpus.
\newblock In {\em ICASSP}, 2003.

\bibitem[\protect\citeauthoryear{Joshi \bgroup \em et al.\egroup
  }{2020}]{drsummarize}
Anirudh Joshi, Namit Katariya, Xavier Amatriain, and Anitha Kannan.
\newblock Dr. summarize: Global summarization of medical dialogue by exploiting
  local structures.
\newblock In {\em Proc. of Findings of EMNLP}, 2020.

\bibitem[\protect\citeauthoryear{Kazi and Kahanda}{2019}]{medical-topic-2}
Nazmul Kazi and Indika Kahanda.
\newblock Automatically generating psychiatric case notes from digital
  transcripts of doctor-patient conversations.
\newblock In {\em Proceedings of the 2nd Clinical Natural Language Processing
  Workshop}, 2019.

\bibitem[\protect\citeauthoryear{Khalman \bgroup \em et al.\egroup
  }{2021}]{forumsum}
Misha Khalman, Yao Zhao, and Mohammad Saleh.
\newblock {F}orum{S}um: A multi-speaker conversation summarization dataset.
\newblock In {\em Findings of the Association for Computational Linguistics:
  EMNLP 2021}, 2021.

\bibitem[\protect\citeauthoryear{Koay \bgroup \em et al.\egroup
  }{2020}]{coling2020-term-koay}
Jia~Jin Koay, Alexander Roustai, Xiaojin Dai, Dillon Burns, Alec Kerrigan, and
  Fei Liu.
\newblock How domain terminology affects meeting summarization performance.
\newblock In {\em Proc. of COLING}, 2020.

\bibitem[\protect\citeauthoryear{Koay \bgroup \em et al.\egroup
  }{2021}]{Koay2021ASA}
Jia~Jin Koay, Alexander Roustai, Xiaojin Dai, and Fei Liu.
\newblock A sliding-window approach to automatic creation of meeting minutes.
\newblock In {\em Proc. of NAACL: Student Research Workshop}, 2021.

\bibitem[\protect\citeauthoryear{Krishna \bgroup \em et al.\egroup
  }{2021}]{soap}
Kundan Krishna, Sopan Khosla, Jeffrey Bigham, and Zachary~C. Lipton.
\newblock Generating {SOAP} notes from doctor-patient conversations using
  modular summarization techniques.
\newblock In {\em Proc. of ACL-IJCNLP}, 2021.

\bibitem[\protect\citeauthoryear{Lei \bgroup \em et al.\egroup
  }{2021}]{speaker}
Yuejie Lei, Yuanmeng Yan, Zhiyuan Zeng, Keqing He, Ximing Zhang, and Weiran Xu.
\newblock Hierarchical speaker-aware sequence-to-sequence model for dialogue
  summarization.
\newblock In {\em Proc. of ICASSP}, 2021.

\bibitem[\protect\citeauthoryear{Lewis \bgroup \em et al.\egroup }{2020}]{bart}
Mike Lewis, Yinhan Liu, Naman Goyal, Marjan Ghazvininejad, Abdelrahman Mohamed,
  Omer Levy, Veselin Stoyanov, and Luke Zettlemoyer.
\newblock {BART}: Denoising sequence-to-sequence pre-training for natural
  language generation, translation, and comprehension.
\newblock In {\em Proc. of ACL}, 2020.

\bibitem[\protect\citeauthoryear{Li \bgroup \em et al.\egroup
  }{2019}]{acl2019-limanling}
Manling Li, Lingyu Zhang, Heng Ji, and Richard~J. Radke.
\newblock Keep meeting summaries on topic: Abstractive multi-modal meeting
  summarization.
\newblock In {\em Proc. of ACL}, 2019.

\bibitem[\protect\citeauthoryear{Lin \bgroup \em et al.\egroup }{2021}]{csds}
Haitao Lin, Liqun Ma, Junnan Zhu, Lu~Xiang, Yu~Zhou, Jiajun Zhang, and
  Chengqing Zong.
\newblock {CSDS}: A fine-grained {C}hinese dataset for customer service
  dialogue summarization.
\newblock In {\em Proc. of EMNLP}, 2021.

\bibitem[\protect\citeauthoryear{Lin}{2004}]{rouge}
Chin-Yew Lin.
\newblock {ROUGE}: A package for automatic evaluation of summaries.
\newblock In {\em Text Summarization Branches Out}, 2004.

\bibitem[\protect\citeauthoryear{Liu and
  Chen}{2021}]{liu-chen-2021-controllable}
Zhengyuan Liu and Nancy Chen.
\newblock Controllable neural dialogue summarization with personal named entity
  planning.
\newblock In {\em Proc. of EMNLP}, 2021.

\bibitem[\protect\citeauthoryear{Liu \bgroup \em et al.\egroup
  }{2019a}]{cs-kdd}
Chunyi Liu, Peng Wang, Jiang Xu, Zang Li, and Jieping Ye.
\newblock Automatic dialogue summary generation for customer service.
\newblock In {\em Proc. of KDD}, 2019.

\bibitem[\protect\citeauthoryear{Liu \bgroup \em et al.\egroup
  }{2019b}]{medical-topic-1}
Zhengyuan Liu, A.~Ng, Sheldon Lee~Shao Guang, AiTi Aw, and Nancy~F. Chen.
\newblock Topic-aware pointer-generator networks for summarizing spoken
  conversations.
\newblock {\em 2019 IEEE Automatic Speech Recognition and Understanding
  Workshop (ASRU)}, 2019.

\bibitem[\protect\citeauthoryear{Liu \bgroup \em et al.\egroup
  }{2021}]{CoreferenceAwareDS}
Zhengyuan Liu, Ke~Shi, and Nancy~F. Chen.
\newblock Coreference-aware dialogue summarization.
\newblock In {\em SIGDIAL}, 2021.

\bibitem[\protect\citeauthoryear{Loza \bgroup \em et al.\egroup
  }{2014}]{email-data-3}
Vanessa Loza, Shibamouli Lahiri, Rada Mihalcea, and Po-Hsiang Lai.
\newblock Building a dataset for summarization and keyword extraction from
  emails.
\newblock In {\em Proc. of LREC}, 2014.

\bibitem[\protect\citeauthoryear{Malykh \bgroup \em et al.\egroup
  }{2020}]{coling2020-titles-malykh}
Valentin Malykh, Konstantin Chernis, Ekaterina Artemova, and Irina
  Piontkovskaya.
\newblock {S}um{T}itles: a summarization dataset with low extractiveness.
\newblock In {\em Proc. of COLING}, 2020.

\bibitem[\protect\citeauthoryear{Mehnaz \bgroup \em et al.\egroup
  }{2021}]{gupshup}
Laiba Mehnaz, Debanjan Mahata, Rakesh Gosangi, Uma~Sushmitha Gunturi, Riya
  Jain, Gauri Gupta, Amardeep Kumar, Isabelle~G. Lee, Anish Acharya, and
  Rajiv~Ratn Shah.
\newblock {G}up{S}hup: Summarizing open-domain code-switched conversations.
\newblock In {\em Proc. of the EMNLP}, 2021.

\bibitem[\protect\citeauthoryear{Mihalcea and Tarau}{2004}]{textrank}
Rada Mihalcea and Paul Tarau.
\newblock {T}ext{R}ank: Bringing order into text.
\newblock In {\em Proc. of EMNLP}, 2004.

\bibitem[\protect\citeauthoryear{Mukherjee \bgroup \em et al.\egroup
  }{2020}]{email-todo-new}
Sudipto Mukherjee, Subhabrata Mukherjee, Marcello Hasegawa, Ahmed
  Hassan~Awadallah, and Ryen White.
\newblock Smart to-do: Automatic generation of to-do items from emails.
\newblock In {\em Proc. of ACL}, 2020.

\bibitem[\protect\citeauthoryear{Nallapati \bgroup \em et al.\egroup
  }{2017}]{aaai2017-summarurnn-nallapati}
Ramesh Nallapati, Feifei Zhai, and Bowen Zhou.
\newblock Summarunner: {A} recurrent neural network based sequence model for
  extractive summarization of documents.
\newblock In {\em Proc. of AAAI}, 2017.

\bibitem[\protect\citeauthoryear{Narayan \bgroup \em et al.\egroup
  }{2021}]{frost}
Shashi Narayan, Yao Zhao, Joshua Maynez, Gon{\c{c}}alo Sim{\~o}es, Vitaly
  Nikolaev, and Ryan McDonald.
\newblock Planning with learned entity prompts for abstractive summarization.
\newblock {\em Transactions of ACL}, 2021.

\bibitem[\protect\citeauthoryear{Paice}{1990}]{old1}
Chris~D Paice.
\newblock Constructing literature abstracts by computer: techniques and
  prospects.
\newblock {\em Information Processing \& Management}, 1990.

\bibitem[\protect\citeauthoryear{Rameshkumar and Bailey}{2020}]{crd3}
Revanth Rameshkumar and Peter Bailey.
\newblock Storytelling with dialogue: {A} {Critical Role Dungeons and Dragons
  Dataset}.
\newblock In {\em Proc. of ACL}, 2020.

\bibitem[\protect\citeauthoryear{See \bgroup \em et al.\egroup }{2017}]{pgn}
Abigail See, Peter~J. Liu, and Christopher~D. Manning.
\newblock Get to the point: Summarization with pointer-generator networks.
\newblock In {\em Proc. of ACL}, 2017.

\bibitem[\protect\citeauthoryear{Shang \bgroup \em et al.\egroup
  }{2018}]{acl2018-shang}
Guokan Shang, Wensi Ding, Zekun Zhang, Antoine Tixier, Polykarpos Meladianos,
  Michalis Vazirgiannis, and Jean-Pierre Lorr{\'e}.
\newblock Unsupervised abstractive meeting summarization with multi-sentence
  compression and budgeted submodular maximization.
\newblock In {\em Proc. of ACL}, 2018.

\bibitem[\protect\citeauthoryear{Song \bgroup \em et al.\egroup
  }{2020}]{coling2020-medical-song}
Yan Song, Yuanhe Tian, Nan Wang, and Fei Xia.
\newblock Summarizing medical conversations via identifying important
  utterances.
\newblock In {\em Proc. of COLING}, 2020.

\bibitem[\protect\citeauthoryear{Tang \bgroup \em et al.\egroup
  }{2021a}]{tang2021investigating}
Xiangru Tang, Alexander~R Fabbri, Ziming Mao, Griffin Adams, Borui Wang, Haoran
  Li, Yashar Mehdad, and Dragomir Radev.
\newblock Investigating crowdsourcing protocols for evaluating the factual
  consistency of summaries.
\newblock {\em arXiv preprint arXiv:2109.09195}, 2021.

\bibitem[\protect\citeauthoryear{Tang \bgroup \em et al.\egroup
  }{2021b}]{Tang2021CONFITTF}
Xiangru Tang, Arjun Nair, Borui Wang, Bingyao Wang, Jai Desai, Aaron Wade,
  Haoran Li, Asli Celikyilmaz, Yashar Mehdad, and Dragomir Radev.
\newblock Confit: Toward faithful dialogue summarization with
  linguistically-informed contrastive fine-tuning.
\newblock {\em ArXiv}, 2021.

\bibitem[\protect\citeauthoryear{Ulrich \bgroup \em et al.\egroup
  }{2008}]{email-data-2}
Jan Ulrich, Gabriel Murray, and Giuseppe Carenini.
\newblock A publicly available annotated corpus for supervised email
  summarization.
\newblock In {\em Proc. of AAAI email workshop}, 2008.

\bibitem[\protect\citeauthoryear{Vaswani \bgroup \em et al.\egroup
  }{2017}]{transformer}
Ashish Vaswani, Noam Shazeer, Niki Parmar, Jakob Uszkoreit, Llion Jones,
  Aidan~N. Gomez, Lukasz Kaiser, and Illia Polosukhin.
\newblock Attention is all you need.
\newblock In {\em Proc. of NeuIPS}, 2017.

\bibitem[\protect\citeauthoryear{Wang \bgroup \em et al.\egroup
  }{2022}]{wang2022survey}
Jiaan Wang, Fandong Meng, Duo Zheng, Yunlong Liang, Zhixu Li, Jianfeng Qu, and
  Jie Zhou.
\newblock A survey on cross-lingual summarization.
\newblock {\em arXiv preprint arXiv:2203.12515}, 2022.

\bibitem[\protect\citeauthoryear{Wu \bgroup \em et al.\egroup }{2021}]{cods}
Chien-Sheng Wu, Linqing Liu, Wenhao Liu, Pontus Stenetorp, and Caiming Xiong.
\newblock Controllable abstractive dialogue summarization with sketch
  supervision.
\newblock In {\em Proc. of Findings of ACL-IJCNLP}, 2021.

\bibitem[\protect\citeauthoryear{Yu \bgroup \em et al.\egroup
  }{2021}]{AdaptSum}
Tiezheng Yu, Zihan Liu, and Pascale Fung.
\newblock {A}dapt{S}um: Towards low-resource domain adaptation for abstractive
  summarization.
\newblock In {\em Proc. of ACL}, 2021.

\bibitem[\protect\citeauthoryear{Zhang \bgroup \em et al.\egroup
  }{2020}]{dialogpt}
Yizhe Zhang, Siqi Sun, Michel Galley, Yen-Chun Chen, Chris Brockett, Xiang Gao,
  Jianfeng Gao, Jingjing Liu, and Bill Dolan.
\newblock {DIALOGPT} : Large-scale generative pre-training for conversational
  response generation.
\newblock In {\em Proc. of ACL}, 2020.

\bibitem[\protect\citeauthoryear{Zhang \bgroup \em et al.\egroup
  }{2021a}]{emailsum}
Shiyue Zhang, Asli Celikyilmaz, Jianfeng Gao, and Mohit Bansal.
\newblock {E}mail{S}um: Abstractive email thread summarization.
\newblock In {\em Proc. of ACL-IJCNLP}, 2021.

\bibitem[\protect\citeauthoryear{Zhang \bgroup \em et al.\egroup
  }{2021b}]{cs-tat}
Xinyuan Zhang, Ruiyi Zhang, M.~Zaheer, and Amr Ahmed.
\newblock Unsupervised abstractive dialogue summarization for tete-a-tetes.
\newblock In {\em Proc. of AAAI}, 2021.

\bibitem[\protect\citeauthoryear{Zhang \bgroup \em et al.\egroup
  }{2021c}]{zhang-etal-2021-exploratory-study}
Yusen Zhang, Ansong Ni, Tao Yu, Rui Zhang, Chenguang Zhu, Budhaditya Deb, Asli
  Celikyilmaz, Ahmed~Hassan Awadallah, and Dragomir Radev.
\newblock An exploratory study on long dialogue summarization: What works and
  what{'}s next.
\newblock In {\em Proc. of Findings of EMNLP}, 2021.

\bibitem[\protect\citeauthoryear{Zhao \bgroup \em et al.\egroup
  }{2020}]{coling2020-topic-zhao}
Lulu Zhao, Weiran Xu, and Jun Guo.
\newblock Improving abstractive dialogue summarization with graph structures
  and topic words.
\newblock In {\em Proc. of COLING}, 2020.

\bibitem[\protect\citeauthoryear{Zhao \bgroup \em et al.\egroup
  }{2021}]{todsum}
Lulu Zhao, Fujia Zheng, Keqing He, Weihao Zeng, Yuejie Lei, Huixing Jiang, Wei
  Wu, Weiran Xu, Jun Guo, and Fanyu Meng.
\newblock Todsum: Task-oriented dialogue summarization with state tracking.
\newblock {\em arXiv}, 2021.

\bibitem[\protect\citeauthoryear{Zhong \bgroup \em et al.\egroup
  }{2021}]{qmsum}
Ming Zhong, Da~Yin, Tao Yu, Ahmad Zaidi, Mutethia Mutuma, Rahul Jha,
  Ahmed~Hassan Awadallah, Asli Celikyilmaz, Yang Liu, Xipeng Qiu, and Dragomir
  Radev.
\newblock {QMS}um: A new benchmark for query-based multi-domain meeting
  summarization.
\newblock In {\em Proc. of ACL}, 2021.

\bibitem[\protect\citeauthoryear{Zhong \bgroup \em et al.\egroup
  }{2022}]{DialogLM}
Ming Zhong, Yang Liu, Yichong Xu, Chenguang Zhu, and Michael Zeng.
\newblock Dialoglm: Pre-trained model for long dialogue understanding and
  summarization.
\newblock {\em Proc. of AAAI}, 2022.

\bibitem[\protect\citeauthoryear{Zhu \bgroup \em et al.\egroup
  }{2020}]{emnlp2020-hmnet-zhu}
Chenguang Zhu, Ruochen Xu, Michael Zeng, and Xuedong Huang.
\newblock A hierarchical network for abstractive meeting summarization with
  cross-domain pretraining.
\newblock In {\em Proc. of Findings of EMNLP}, 2020.

\bibitem[\protect\citeauthoryear{Zhu \bgroup \em et al.\egroup
  }{2021}]{naacl2021-media-zhu}
Chenguang Zhu, Yang Liu, Jie Mei, and Michael Zeng.
\newblock {M}edia{S}um: A large-scale media interview dataset for dialogue
  summarization.
\newblock In {\em Proc. of NAACL}, 2021.

\bibitem[\protect\citeauthoryear{Zou \bgroup \em et al.\egroup
  }{2021a}]{cs-unsupervised-zou}
Yicheng Zou, Jun Lin, Lujun Zhao, Yangyang Kang, Zhuoren Jiang, Changlong Sun,
  Qi~Zhang, Xuanjing Huang, and Xiaozhong Liu.
\newblock Unsupervised summarization for chat logs with topic-oriented ranking
  and context-aware auto-encoders.
\newblock In {\em Proc. of AAAI}, 2021.

\bibitem[\protect\citeauthoryear{Zou \bgroup \em et al.\egroup
  }{2021b}]{cs-supervised-zou}
Yicheng Zou, Lujun Zhao, Yangyang Kang, Jun Lin, Minlong Peng, Zhuoren Jiang,
  Changlong Sun, Qi~Zhang, Xuanjing Huang, and Xiaozhong Liu.
\newblock Topic-oriented spoken dialogue summarization for customer service
  with saliency-aware topic modeling.
\newblock In {\em Proc. of AAAI}, 2021.

\end{thebibliography}

\end{document}